\documentclass{article}

 \usepackage{graphicx}
 \usepackage{amsmath}
 \usepackage[preprint]{neurips_2025}

\usepackage[utf8]{inputenc} 
\usepackage[T1]{fontenc}    
\usepackage{hyperref}       
\usepackage{url}            
\usepackage{booktabs}       
\usepackage{amsfonts}       
\usepackage{nicefrac}       
\usepackage{microtype}      
\usepackage{xcolor}         

\title{Sparse Crosscoders for diffing MoEs and Dense models}

\author{
  \begin{tabular}{ccc}
    \textbf{Marmik Chaudhari}$^{1}$ & \textbf{Nishkal Hundia}$^{2}$ & \textbf{Idhant Gulati}$^{1}$ \\
    \texttt{marmik@berkeley.edu} & \texttt{nhundia@umd.edu} & \texttt{idhant@berkeley.edu} \\
    \multicolumn{3}{c}{\vspace{0.5em}} \\
    \multicolumn{3}{c}{\normalfont $^1$University of California, Berkeley \quad $^2$University of Maryland} \\
  \end{tabular}
}

\begin{document}

\maketitle

\begin{abstract}

Mixture of Experts (MoE) achieve parameter-efficient scaling through sparse expert routing, yet their internal representations remain poorly understood compared to dense models. We present a systematic comparison of MoE and dense model internals using crosscoders, a variant of sparse autoencoders, that jointly models multiple activation spaces. We train 5-layer dense and MoEs (equal active parameters) on 1B tokens across code, scientific text, and english stories. Using BatchTopK crosscoders with explicitly designated shared features, we achieve $\sim$ 87\% fractional variance explained and uncover concrete differences in feature organization. The MoE learns significantly fewer unique features compared to the dense model. MoE-specific features also exhibit higher activation density than shared features, whereas dense-specific features show lower density. Our analysis reveals that MoEs develop more specialized, focused representations while dense models distribute information across broader, more general-purpose features.

\end{abstract}

\section{Introduction}
Mixture of Experts (MoEs) \citep{moe-shazeer} are a class of neural networks that route each input through only a small subset of specialized "experts", rather than activating all parameters. This sparse activation pattern allows MoEs to scale model capacity without proportionally increasing computational cost during inference. Because of these efficiency gains, MoEs have recently gained significant attention in the development of large language models, with many of the state-of-the-art open-source models adopting the architecture (e.g., DeepSeek-V3 \citep{deepseek-v3}, Switch Transformer \citep{switch-transformers-fedus}).

Despite this momentum, there has been relatively little research into their internal structure and how they mechanistically compare to dense models. While dense models have been the subject of extensive interpretability research, from understanding attention patterns \citep{struct-attn-vig} to discovering interpretable features through dictionary learning (\cite{sparseautoencoders-cunningham}; \cite{towards-3monosemanticity-bricken}), there has been comparatively little work exploring how MoEs represent information, how their experts specialize, and how their internal representations differ from those of dense models. Given the fundamental architectural differences between sparse and dense models, it is unclear whether existing intuitions and findings about dense model internals carry over to MoEs.

Understanding a model's internal structure often involves studying its features, the learned representations that capture meaningful patterns in the data. Recent work has shown that features in dense models can be systematically extracted and analyzed using techniques such as sparse autoencoders and probing methods. However, the sparse routing mechanism in MoEs introduces several questions: Do experts develop distinct feature representations? How does the routing strategy influence feature specialization? And how do the overall learned representations compare between architectures with similar capacity but different activation patterns?

In this work, we address these questions through a systematic comparison of MoE and dense model internals. Specifically, we (1) train a 5-layer MoE model and a comparable dense model on the same dataset with matched active parameters, (2) adapt existing dense model interpretability techniques to function with MoEs, (3) systematically analyze and compare the internal representations of both models to identify shared and model-specific features, and (4) investigate how sparsity and expert routing influence feature specialization and diversity. Our results show that, compared to the dense model, the MoE learns fewer but more specific features, suggesting that sparsity encourages localized specialization. These findings provide new insights into how MoEs internally organize information and offer a foundation for future mechanistic interpretability studies in sparse architectures.

\section{Background}

\subsection{Mixture of Experts}
MoEs are neural networks in which only a subset of parameters is activated for each input token. Instead of using a single dense feedforward layer, MoE layers are partitioned into multiple smaller "expert" MLPs. A router network determines which experts to activate for each token, typically selecting the top-$k$ experts based on the token’s representation. This sparse activation mechanism enables MoE models to scale to a much larger number of total parameters than dense models, while maintaining comparable computational cost during inference, since only a fraction of the parameters are used per token.


\subsection{Crosscoders}
Crosscoders \citep{crosscoders-lindsey} extend sparse autoencoders to jointly model two activation spaces, learning shared sparse features $f_i(x)$ that reconstruct both spaces through model-specific decoder weights. We adopt the BatchTopK variant with explicitly designated shared features \citep{overcoming-sparsity-minder}, which enforces hard sparsity constraints and improves interpretability of model-exclusive features.

Formally, given activations $x^A \in \mathbb{R}^{d_A}$ from model $A$ and $x^B \in \mathbb{R}^{d_B}$ from model $B$ on the same input, a crosscoder learns sparse feature activations $\{f_i(x)\}_{i=1}^k$ and decoder vectors $\{W^A_{\text{dec}, i}\}_{i=1}^k$, $\{W^B_{\text{dec}, i}\}_{i=1}^k$ such that

\begin{equation}
\hat{x}^m = \sum_{i=1}^k f_i(x)\, W^m_{\text{dec}, i} \quad \text{for } m \in \{A, B\}
\end{equation}

where $k$ is the number of learned features and $f_i(x)$ is the activation of feature $i$ on input $x$. The training objective combines reconstruction losses with sparsity regularization:
\begin{equation}
\mathcal{L} = \mathbb{E}_x \Bigg[
    \lVert x^A - \hat{x}^A \rVert^2 +
    \lVert x^B - \hat{x}^B \rVert^2 +
    \lambda \sum_{i=1}^k f_i(x)\bigl(\lVert W^A_{\text{dec}, i} \rVert + \lVert W^B_{\text{dec}, i} \rVert \bigr)
\Bigg].
\end{equation}

By using the same sparse features $f_i(x)$ to reconstruct both activation spaces, crosscoders naturally identify which features are shared and which are model-specific. Features can be classified according to the relative magnitudes of their decoder weights across the two models.

\textbf{Fixed shared-feature variant:} To improve interpretability of model-exclusive features, \cite{crosscoders-lindsey} propose designating a subset $S \subset \{1,\dots,k\}$ of features as explicitly shared between models. These features have tied decoder parameters across models and a reduced sparsity penalty relative to the exclusive features $F = \{1,\dots,k\} \setminus S$. The objective becomes
\begin{equation}
\mathcal{L} = \mathbb{E}_x \Bigg[
   \sum_{m \in \{A,B\}} \lVert x^m - \hat{x}^m \rVert^2
   + \lambda_s \sum_{i \in S} f_i(x)\, \lVert W_{\text{dec}, i} \rVert
   + \lambda_f \sum_{i \in F} f_i(x)\, \sum_{m \in \{A,B\}} \lVert W^m_{\text{dec}, i} \rVert
\Bigg],
\end{equation}
where $\lambda_s$ and $\lambda_f$ are the sparsity penalties for the shared and exclusive features respectively. Keeping $\lambda_s / \lambda_f \approx 0.1$--$0.2$ encourages shared features to capture common variance, relieving pressure on exclusive features and leading to more monosemantic, interpretable model-specific latents.

\textbf{BatchTopK crosscoder.} The BatchTopK variant \citep{overcoming-sparsity-minder} replaces continuous L1 penalties with a hard sparsity constraint. For a batch of inputs, it selects the top activations
\[
\operatorname{TopK}\Bigl(f_i(x)\, \bigl(\lVert W^A_{\text{dec}, i}\rVert + \lVert W^B_{\text{dec}, i}\rVert \bigr)\Bigr)
\]
across all features $i$ and sets the remaining activations to zero. This enforces a fixed sparsity budget across the batch, prioritizing features with stronger activations and larger decoder norms.

\section{Methods}

We train 5-layer Dense and Mixture of Experts models on a dataset of around 1 billion tokens. The dataset comprises of equal subsets $(\approx 333 \text{M tokens})$ of Arxiv from RedPajama \citep{red-pajamas-weber}, code from StarCoder \citep{star-coder} and English stories from SimpleStories \citep{SimpleStories-parameterized-synthetic-finke}. Both the models are trained on the standard Cross Entropy loss. The MoE was trained with an additional Switch load balancing loss following \citep{switch-transformers-fedus}. Both models were trained for 2 epochs.

Subsequently we train a crosscoder on the outputs of the third layer from both the models. We then leverage the learned decoder weights of the crosscoder dictionary to compare the models. Although the latent activations $f_j(x)$ are shared between both models, the corresponding decoder vectors, $\mathbf{W}_i^{\mathrm{MoE}}$ and $\mathbf{W}_i^{\mathrm{dense}}$ are specific to each model \citep{overcoming-sparsity-minder}. To check the model-specificity of a particular feature, we compute  the relative difference of decoder latent norms $\Delta_{\mathrm{norm}}$, given by:

\begin{equation}
\Delta_{\mathrm{norm}}(i) = \frac{1}{2} \left( \frac{\lVert \mathbf{W}_i^{\mathrm{dense}} \rVert_2 - \lVert \mathbf{W}_i^{\mathrm{MoE}} \rVert_2}{\max\left( \lVert \mathbf{W}_i^{\mathrm{dense}} \rVert_2, \lVert \mathbf{W}_i^{\mathrm{MoE}} \rVert_2 \right)} + 1 \right)
\end{equation}

$\Delta_{\mathrm{norm}}$ measures how much feature $i$ is specialized to a particular model versus being shared. $\Delta_{\mathrm{norm}}=0.5$ represents feature equally shared between the two models, $\sim 0$ represents feature exclusive to the $\text{MoE}$ and $\sim 1$ represents feature exclusive to the $\text{Dense}$ model.


\section{Results}

\begin{figure}[h!]
    \centering
    \begin{minipage}{0.52\linewidth}
        \centering
        \includegraphics[width=\linewidth]{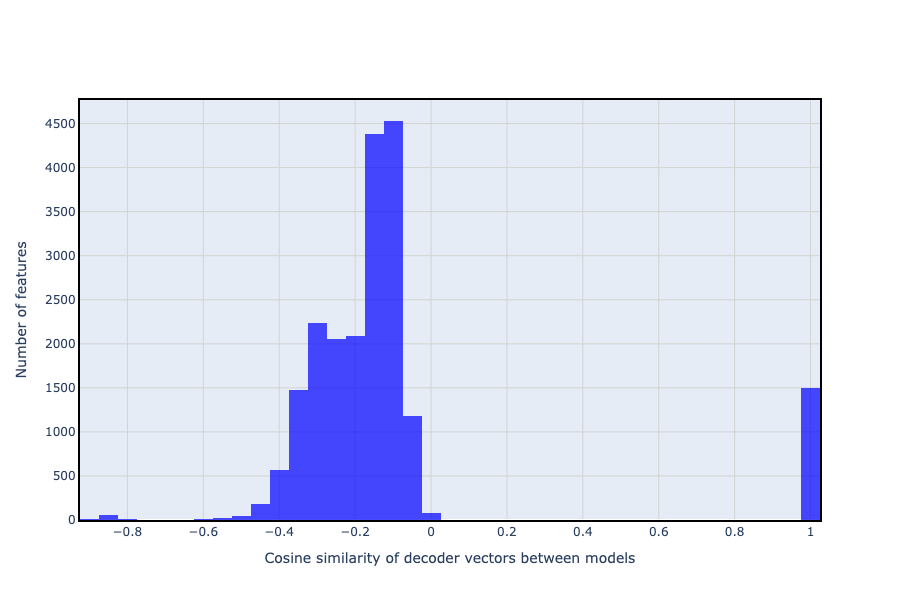}
    \end{minipage}\hfill
    \begin{minipage}{0.48\linewidth}
        \centering
        \includegraphics[width=\linewidth]{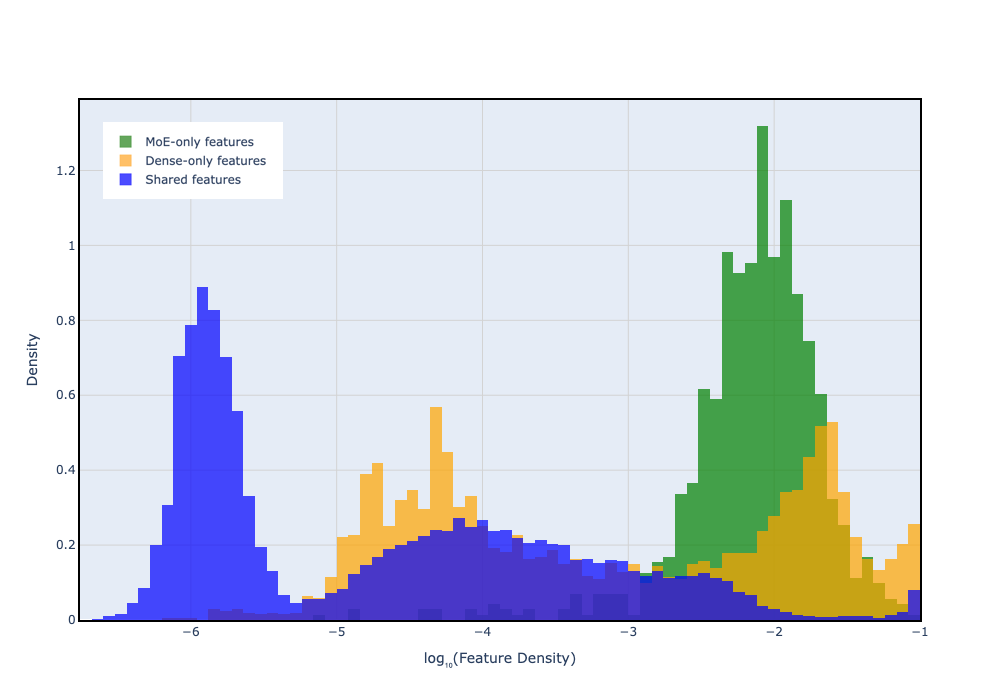}
    \end{minipage}
    \caption{(Left) Comparison between cosine similarity of decoder vectors between the MoE and dense model and (Right) feature densities of the shared, dense and MoE-specific features where x-axis shows the activation frequency of features and y-axis shows the density of features.}
    \label{fig:cosine-density}
\end{figure}

Training a standard crosscoder on the dense and MoE resulted in a large number of features being classified as shared, despite their decoder weights having near-zero cosine similarity as shown in Figure \ref{fig:cosine-density}. This suggests that the standard crosscoder objective overestimates shared structure when the activations of the compared models differ substantially.

\begin{figure}[htbp]
    \centering
    \begin{minipage}{0.48\textwidth}
        \centering
        \includegraphics[width=\linewidth]{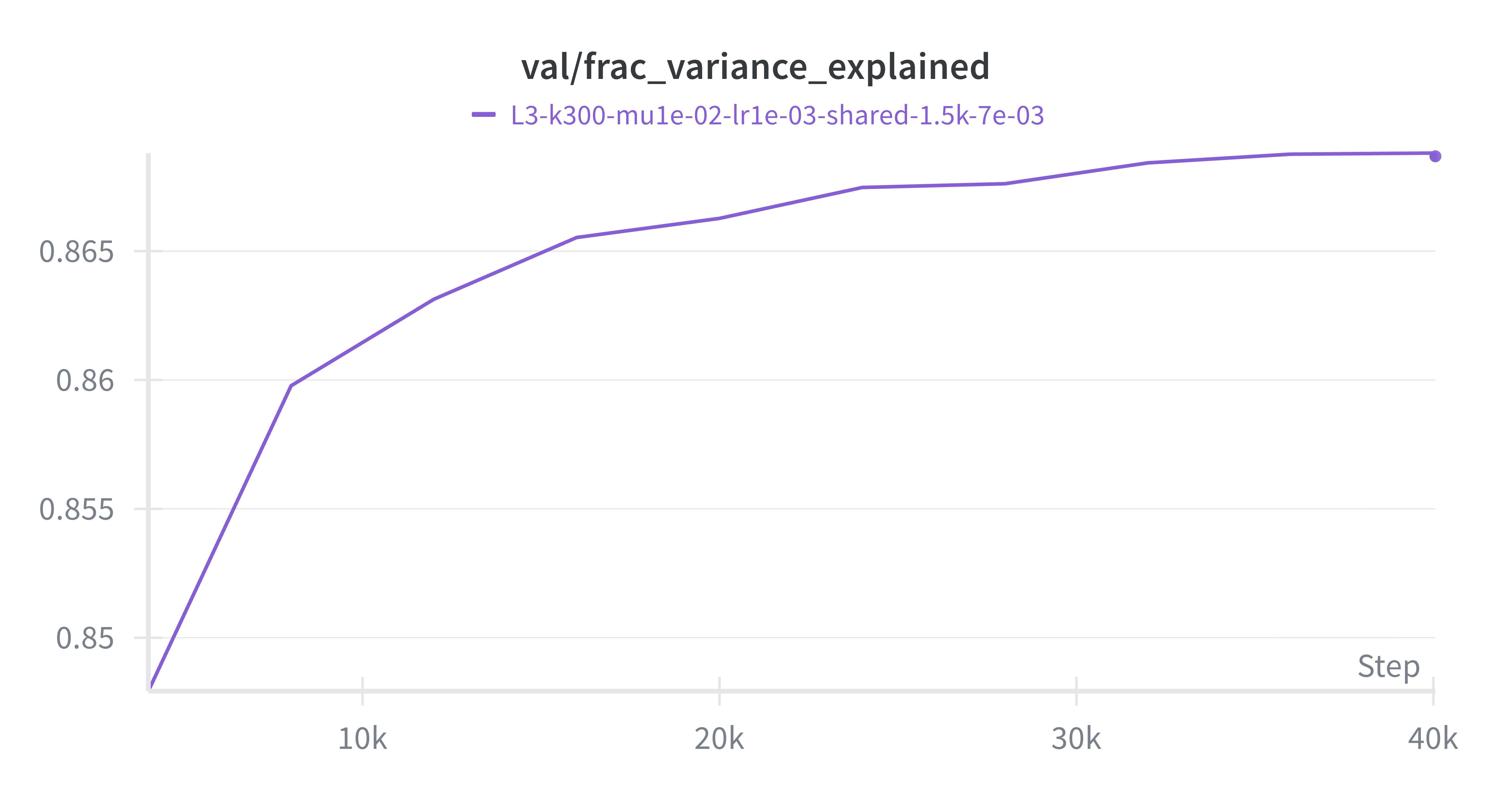}
        \caption{Fractional variance explained of model activations across 40k training steps}
        \label{fig:frac-variance}
    \end{minipage}
    \hfill
    \begin{minipage}{0.48\textwidth}
        \centering
        \includegraphics[width=\linewidth]{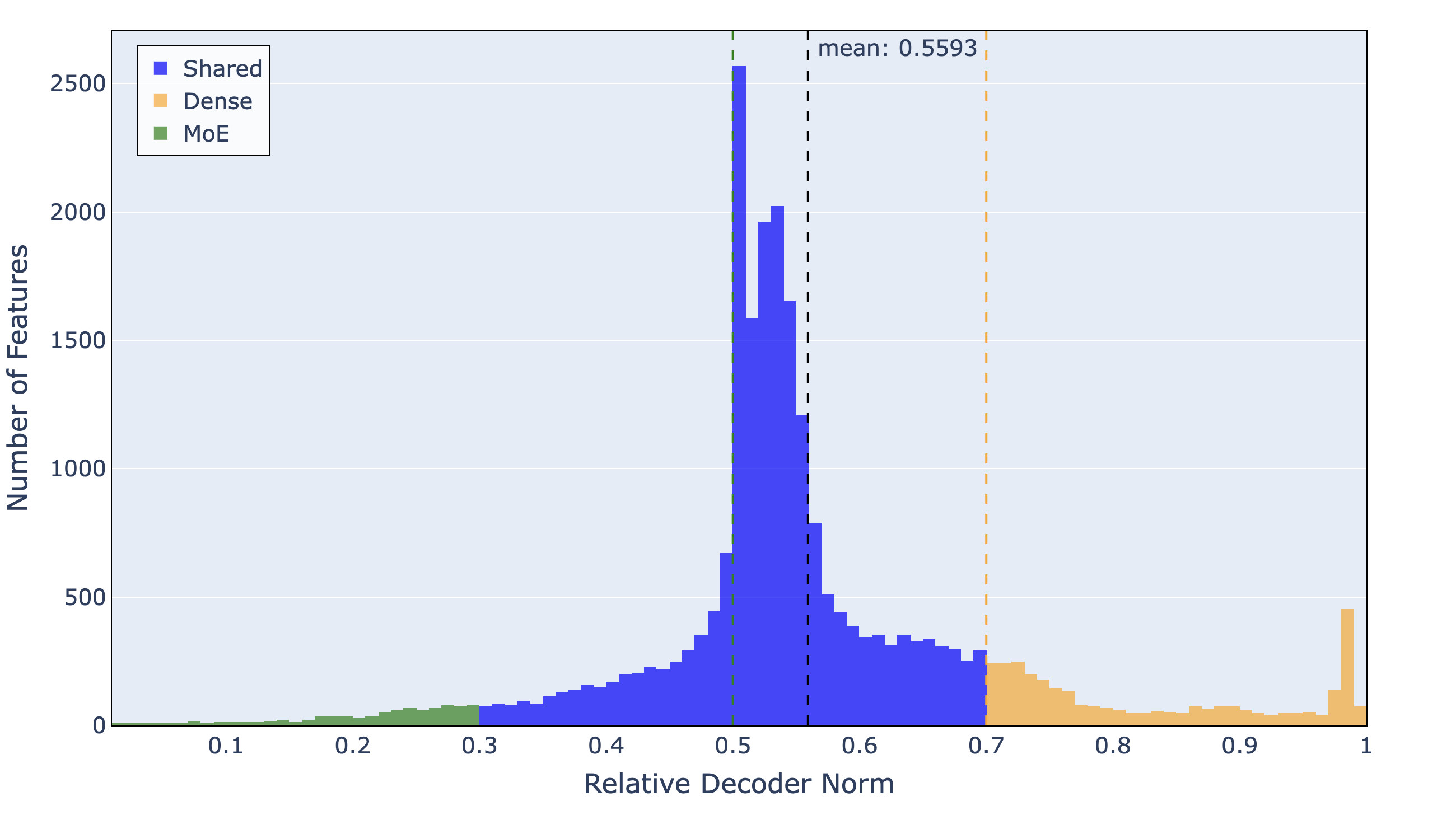}
        \caption{Relative difference of decoder norm vectors for features in different models. MoE specific features on the left $(<0.3)$ and Dense specific features on the right $(>0.7)$.}
        \label{fig:decoder-dist}
    \end{minipage}
\end{figure}

To address this, we applied the variation described earlier, which explicitly designates a subset of features as shared. We experimented with different values of $\lambda_s$ and the number of fixed shared features. Interestingly, the empirical ratio of $\lambda_s/\lambda_f = 0.1$-$0.2$ suggested by prior work did not transfer well to our setting, leading to high $\ell_0$ loss and poor reconstructions. We hypothesize that this discrepancy arises because prior experiments focused on comparing a model with a fine-tuned version of itself, whereas in our case the dense and MoE models were trained independently from scratch. This increased divergence between activation spaces likely necessitates a stronger shared-feature regularization term. Consequently, we found that a higher ratio of approximately $0.7$ was required for effectively distinguishing between model specific features.

Combining this variation with the BatchTopK encoder produced our best-performing crosscoder, achieving a high fractional variance explained of model activations, specifically $\sim 87\%$ over 40K training steps as shown in Figure \ref{fig:frac-variance}. 



We observed that the crosscoder identified a significantly higher number of unique features for the dense model as compared to the MoE as stated in Table 1. We also observed that MoE-only features have higher feature densities than the shared features while the dense-only features have lower feature densities than the shared features. This differs from crosscoders trained on base and fine-tuned models where both model-specific features have higher feature densities than the shared features as noted in \citep{crosscoders-lindsey}.

\begin{table}[h]
\centering
\caption{Distribution of shared, dense and MoE-specific features by $\Delta_{\text{norm}}$}
\begin{tabular}{lcc}
\toprule
Name & $\Delta_{\text{norm}}$ & Count \\
\midrule{MoE-only features} & 0.0--0.3 & 910 \\
{Dense-only features} & 0.7--1 & 3,226 \\
{Shared features} & 0.3--0.7 & 18,940 \\
\bottomrule
\end{tabular}
\end{table}

We observe high cosine similarity $\sim 1$ for the budgeted shared features but the other shared features in the $\Delta_{\mathrm{norm}}$ range between $0.3$ and $0.7$ that the crosscoder found do not exhibit high cosine similarity with some of the features in the range having exactly opposite direction, cosine similarity $\approx -1$

We do not observe a clear trimodal structure of features that corresponds to MoE-only, shared and Dense-only features as shown in Figure \ref{fig:decoder-dist} which is in contrast to the usual trimodal distribution seen in case of crosscoders for diffing base and finetuned models as seen in \citep{overcoming-sparsity-minder}

\section{Conclusion}

We presented a systematic comparison of MoE and dense model activations using crosscoders, achieving $\approx 87\%$ fractional variance of activations explained. The crosscoder learns significantly fewer MoE-specific features compared to the dense-specific features. Furthermore, the MoE features have high feature densities compared to the shared features while the dense features have relatively lower feature densities. Our work demonstrates that crosscoders can be applied to some extent beyond fine-tuning analysis to understand architectural differences. However, significant work remains to modify crosscoders to better capture differences in activations across structurally distinct models like MoEs and dense models. Future research should focus on conducting qualitative analysis of the discovered features to validate their semantic meaningfulness and relevance.

\bibliographystyle{apalike}  
\bibliography{references}




\end{document}